\documentclass[journal]{IEEEtran}

\usepackage[defaultbib]{bibtopic}
\usepackage[procnumbered,ruled]{algorithm2e}
\usepackage{blindtext}
\usepackage{graphicx}
\usepackage{multicol}

\usepackage{algorithmic}
\usepackage{float}
\makeatletter
\adddialect\l@ENGLISH\l@english
\makeatother

\usepackage{xcolor}
\usepackage{colortbl,hhline}

\ifCLASSINFOpdf

\else

\fi
\usepackage{algorithmic}
\usepackage{float}
\makeatletter
\adddialect\l@ENGLISH\l@english
\makeatother

\usepackage{xcolor}
\usepackage{colortbl,hhline}

\ifCLASSINFOpdf

\else

\fi

\usepackage[utf8]{inputenc}
\usepackage[T1]{fontenc}
\usepackage[english]{babel}
\usepackage{bbm}
\usepackage{amsmath,amsfonts,amssymb}
\usepackage[justification=centering]{caption}

\usepackage[utf8]{inputenc}
\usepackage[T1]{fontenc}
\usepackage[english]{babel}
\usepackage{bbm}
\usepackage{amsmath,amsfonts,amssymb}
\usepackage[justification=centering]{caption}
\begin{document}
%
\title{Dimensionality reduction with missing values imputation}

\author{\IEEEauthorblockN{Rania Mkhinini Gahar$^1$, Olfa Arfaoui$^2$, Minyar Sassi Hidri$^3$, Nejib Ben-Hadj Alouane$^4$}\\
\IEEEauthorblockA{$^{1,2,3,4}$University of Tunis El Manar\\$^{1,2,3,4}$National Engineering School of Tunis\\
$^{1,2,3,4}$BP. 37, Le Belv\`ed\`ere 1002, Tunis, Tunisia\\
$^{1,4}$ OASIS Laboratory\\
$^{2,3}$ LR-SITI Laboratory\\
$^{1,2,3}$ Email:\{rania.mkhininigahar, olfa.arfaoui, minyar.sassi\}@enit.rnu.tn\\
$^{4}$ Email: nejib bha@yahoo.com}}

\maketitle

\begin{abstract}
In this study, we propose a new statical approach for high-dimensionality reduction of heterogenous data that limits the curse of dimensionality and deals with missing values. To handle these latter, we propose to use the Random Forest imputation's method. The main purpose here is to extract useful information and so reducing the search space to facilitate the data exploration process. Several illustrative numeric examples, using data coming from publicly available machine learning repositories are also included. The experimental
component of the study shows the efficiency of the proposed analytical approach.
\end{abstract}

 \section{Introduction}
  For about thirty years, data analysis methods have largely demonstrated their effectiveness in the processing of data in many fields. Data reduction  is one of these methods and  part of the descriptive (or exploratory) statistics. It tries to summarize a sample of data using graphs or numerical characteristics.\\
The main interpretation of data reduction is reducing the number of dimensions. This implies that data reduction is part of the multivariate exploratory statistics which seek to reduce the number of data dimensions by extracting a number of factors, dimensions, clusters, etc., which explain the dispersion of (multidimensional) data.

In the case where it is certain that data (which is in most cases voluminous) contains useful information, the reduction methods can be  applied in order to have a synthetic view and so to make easy the exploration process. The issue comes down to a problem of data structuring and knowledge extraction.

In this context, the simultaneous introduction of both quantitative and qualitative variables in a dataset for analysis becomes a frequent problematic when we talk about reducing heterogenous data.

In the context of multidimensional data reduction, it is very frequent for the variables to be of a mixed nature (quantitative and qualitative variables). The usual processing consists in either putting the qualitative variables (resp. quantitative) into illustrative elements in a Principal Component Analysis (PCA)\cite{tenenhaus} (resp. Multiple Correspondence Analysis (MCA)\cite{mca}), or discretizing the quantitative variables into qualitative variables according to an MCA. Which very often introduces a bias due to the choice of the number of classes and their amplitudes equal or different, and which causes a loss of information.

Many researchers are interested in this issue and have proposed methods that deal simultaneously with the two types of variables into active elements: the PCA with indicators introduced in \cite{tenenhaus} and more recently the Factor Analysis of Mixed Data (FAMD) proposed in \cite{pagesbis}.
In fact, the methods of reducing the data representation space size  can be processed either by the extraction or selection of the attributes.
The extraction of the attributes transforms the initial attribute space into a new space formed by the linear or non-linear combination of the initial attributes. Contrariwise, the selection of attributes selects the most relevant attributes according to a given criterion.

In this direction, and to answer the previous questions, we propose in this paper to prove that the high-dimensionality reduction is really a serious challenge, and this, by suggesting a new statistical approach that reduce the huge amount of heterogeneous data without affecting their value.
\section{Review of related work}
The dimension reduction consists in transforming data represented in a large space into a representation in a space of smaller dimension \cite{ideal}. In many fields, the dimension reduction is considered as a very important step because it facilitates the classification, visualization or compression of large data. The main goal here is to limit the effect of issues caused by the high dimensionality data \cite{curse}.

Recently, multiple methods for dimension reduction have been proposed \cite{curse, meth2, meth3, meth4, meth5, meth6, olfa, meth8, meth7}. They are able to deal with complex non-linear problems and have often been proposed as an alternative to conventional linear methods such as the PCA which is one of the most widely used methods of analyzing multivariate data. Once we have a table of quantitative data (continuous or discrete) in which $n$ observations (individuals, products, etc.) are described by $p$ variables (descriptors, attributes, measurements, etc.), if $p$ is quite high, it is impossible to understand the structure of the data and the proximity between observations by simply analyzing univariate descriptive statistics or even a correlation matrix.

In the same context, an other alternative appears which is based on the MCA \cite{mca, mca2, mca3}. This method is also called $\mathbf{CATPCA}$ which refers to $\mathbf{CAT}$egorical $\mathbf{PCA}$. In this method, variable modalities will be as far apart as possible from each other. MCA yields two clouds of points : the cloud of individuals and the cloud of categories. It involves the notion of the complete disjunctive table. In principle, the establishment of a complete disjunctive table is very simple. It consists in coding qualitative variables with value $1$ for the observed modality and $0$ for any other modality (sum of rows = number of variables). This technique  allows calculation on qualitative characteristics used for MCA.

In \cite{afc2} and \cite{afc3}, the authors highlighted the Correspondence Factor Analysis method (CFA) which is a factorial method of Multidimensional Descriptive Statistics. Its objective is to analyze the link between two qualitative variables (if there are more than two qualitative variables, they used the MCA). This method is a particular PCA carried out on the profiles associated with the contingency table crossing the two considered variables. In \cite{mfa2}, the Multiple Factor Analysis (MFA) presents an other alternative used to analyze a set of observations described by several groups of variables \cite{mfa1}. The number of variables in each group may differ and the nature of the variables (nominal or quantitative) can vary from one group to other but the variables should be of the same nature in a given group. The analysis derives an integrated picture of the observations and of the relationships between the groups of variables .

The convergence between these previous three methods (PCA, MCA and MFA) provides a solid justification for a methodology which deserves to be denominated in its own right : Factorial Analysis of Mixed Data (FAMD). The properties of the FAMD are studied in detail in \cite{pagesbis} which presents an application on real data. By introducing a table of mixed data in which each variable (quantitative or qualitative) constitutes a group, the quantitative variables are centered-reduced and the qualitative variables are coded as in MCA. The FAMD is a true generalization (in the sense that the PCA and the MCA are special cases).
\section{High-Dimensionality Reduction of heterogenuous data}

In this section we introduce a new approach to reduce high-dimensional heterogenous data. The basic steps of this approach are illustrated in algorithm \ref{alg3}.
\begin{algorithm}[h!]
\caption{High-Dimensionality Reduction OF Heterogenuous Data.}
\label{alg3}
\small{
\begin{algorithmic}[1]
\renewcommand{\algorithmicrequire}{\textbf{Input:}}
\renewcommand{\algorithmicensure}{\textbf{Output:}}
\REQUIRE $X$ ($n\times p$ matrix) : Messy heterogenous data with missing values.
\ENSURE $X'$ ($n \times p'$ matrix) : Cleaned and reduced data
\STATE Splitting the quantitative and the qualitative variables.
\STATE Applying the reduction centering function on the quantitative variables.
\STATE Applying the complete disjunctive function on the qualitative variables.
\STATE Applying the weighting on the indicators.
\STATE Combining the two matrix resulting from step~~3 and step~~5 (combination by columns) and round the values of the new matrix to 3 decimal digits.
\STATE Performing a principal components analysis on the given data matrix and returning the results as an object of class, then checking the rotation.
\STATE Checking the standard deviation of principal components and compute the proportion variance of each component.
\STATE Checking the proportion variance of components that contain more than 90\% of the information.
\end{algorithmic}}
\end{algorithm}
\subsection{Data cleaning and preparation}
Before applying our statistical approach, a data cleaning and preparation step should take place and it consists of: 1) Exploring raw data, 2) Tidying data and 3) Imputing missing values with RadomForest algorithm \cite{missforest1}. We draw attention to the fact that despite the increasing of the  data amount  and the emergence of Big Data, missing data problems remain widespread in statistical problems and require a particular approach. Since our approach aims to reduce this deluge of data, we propose to apply the RandomForest algorithm \cite{missforest1}which is a missing data imputation algorithm for mixed dataset. This algorithm aims to predict individual missing values accurately rather than take random draws from a distribution, so the imputed values may lead to biased parameter estimated in statistical models. A discription of the RandomForest algorithm is illustrated in algorithm \ref{alg1}.
\begin{algorithm}[h!]
\caption{RandomForest algorithm.}
\label{alg1}
\small{
\begin{algorithmic}[1]
\REQUIRE $\mathbf{X}$ : an $n\ \times p$ matrix, stopping criterion $\gamma$
\ENSURE The imputed matrix $\textbf{X}^{imp}$
\STATE Make initial guess for missing values
\STATE $\textbf{k}$$\leftarrow$ vector of sorted indices of columns in $\mathbf{X}$\\
with respect to increasing amount of missing values
\WHILE {not $\gamma$}
\STATE $\textbf{X}^{imp}_{old}$ $\leftarrow$ store previously imputed matrix
\FOR {$s$ in $\textbf{k}$}
\STATE Fit a random forest: $y_{obs}^{(s)} \sim  x_{obs}^{(s)}$
\STATE Predict $y_{mis}^{(s)}$ using $x_{mis}^{(s)}$
\STATE $\textbf{X}^{imp}_{new}$ $\leftarrow$ update imputed matrix using predicted $y_{mis}^{(s)}$ ;
\ENDFOR
\STATE Update $\gamma$
\ENDWHILE
\end{algorithmic}
}
\end{algorithm}

The stopping criterion $\gamma$ is reached as soon as the difference between the matrix of newly imputed data and the previous one increases for the first time.

The difference of the set of continuous variables $N$ is defined as shown in the equation \eqref{quant}.
\begin{equation}
\label{quant}
\bigtriangleup_{N}=\frac{\sum_{j\in N}(\textbf{X}^{imp}_{new}-\textbf{X}^{imp}_{old})^{2}}{\sum_{j\in N}(\textbf{X}^{imp}_{new})^{2}}
\end{equation}

In the case of qualitative variables $F$, the difference will be as described in the equation \eqref{qual}.
\begin{equation}
\label{qual}
\bigtriangleup_{F}=\frac{\sum_{j\in F}\sum_{i=1}^{n}\mathbbm{1}_{{\textbf{X}^{imp}_{new}\neq\textbf{X}^{imp}_{old}}}}{\#NA}
\end{equation}

where \#NA is the number of missing values in the categorical variables.

To check the imputation error, we have used two concepts : \textit{NRMSE} (NoRmalized Mean Squared Error) \cite{nmse} and \textit{PFC} (Proportion of Falsely Classified) \cite{pfc}. The \textit{NRMSE} is used to represent error derived from imputing continuous values whereas \textit{PFC} is used to represent error derived from imputing categorical values.

\subsection{Splitting the quantitative and the qualitative variables}
Mixed variables may have different distributions in a dataset:
\begin{enumerate}
\item All quantitative variables precede all qualitative variables.
\item All qualitative variables precede all quantitative variables.
\item Random distribution of different types of variables.
\end{enumerate}
\subsection{Applying the centering reduction function on the quantitative variables}
In this step, we apply a centering reduction function on the quantitative variables. The reduced center variable or z-score allows to indicate how many standard deviations, above or below the mean, is a sample of data series. To find the reduced centered variable of an item in the sample, we will need to find the mean, variance and standard deviation of the sample. Then we will have to differentiate between the value of this element and the sample mean and then divide that result by the standard deviation of the same sample. If we follow the following steps, we will see that calculating the reduced center variable is not as difficult as it sounds. Certainly, this makes a lot of calculations, but they are simple.
\subsection{Applying the complete disjunctive function on the qualitative variables}
This step deals with the application of the complete disjunctive table \cite{disj} on the qualitative variables. A complete disjunctive table is a type of qualitative data representation used in data analysis. In this table, a qualitative variable with \textbf{K} modalities is replaced by \textbf{K} binary variables, each corresponding to one of the modalities.
\subsection{Applying the weighting on the indicators}
The weighting of the data consists in assigning a weight to each individual in the initial dataset. The primary objective  is to correct the representativeness of the initial dataset according to certain key variables in order to facilitate the extrapolation of the results \cite{ponderation}.
\subsection{Data transformed to PCA}
In this step, we combine the resulted table of the centering and reduction function and the complete disjunctive table. This combination is called $combination~by~columns$. After that, we round the values of this combination to 3 decimal digits.
\subsection{Performance of the PCA and checking of the rotation}
Here, we apply the PCA method and we select the principal components. The rotation measure provides the principal component loading. Each column of rotation matrix contains the principal component loading vector. This is the most important measure we should be interested in. The rotation returns principal components. Absolutely, in a dataset, the maximum number of principal component loadings is a minimum of $(n-1, p)$.
\subsection{Standard deviation and variance proportion of each component}
We aim to find the components which explain the maximum variance. This is because we want to retain as much information as possible using these components. So, higher is the explained variance, higher will be the information contained in those components.
\section{Computational results}
In this section, we present the experimental results of our large-scale heterogenous data reduction technique. The experiments have been yielded  on two high-dimensional datasets obtained from the University of California at Irvine (UCI) Machine Learning Repository \cite{uci}.
\subsection{Data of experimentation}
The used datasets are the following:
\begin{itemize}
\item Internet Advertisements Data: this dataset was constructed by Nicholas Kushmerick \cite{internetData} and represents a set of possible advertisements on Internet pages. The features encode the geometry of the image (if available) as well as phrases occuring in the URL, the image's URL and alt text, the anchor text, and words occuring near the anchor text. The task is to predict whether an image is an advertisement ("ad") or not ("nonad"). This dataset consists of 10000 quantitative attributes with 1500 instances.
\item Amazon Commerce Reviews data: this dataset is derived from the customers reviews in Amazon Commerce Website for authorship identification. Most previous studies conducted the identification experiments for two to ten authors. Data is in raw form and contains columns of data for real variables. It is created by ZhiLiu et al. \cite{amazonData1} and cited as example in \cite{amazonData2}. We use a version of this dataset with 3279 instances and 1558 mixed attributes. .
\end{itemize}

The basic information of these benchmark datasets is illustrated in table \ref{details}.

\begin{table}[h!]
\begin{center}
\scriptsize{
\centering\caption{Details of the used datasets.}\label{details}
\begin{tabular}[t]{c c c c}
    \hline
    \textit{Datasets}&\textit{\#Attributes}&\textit{\#Attributes Type}&\textit{\#Instances}\\
     \hline
Internet Advertisements  & 1559&Mixed & 3279\\
\hline
Amazon Commerce Reviews & 10000& Quantitative& 1500\\
\hline
  \end{tabular}}
  \end{center}
   \end{table}

\subsection{Results and evaluation}
\begin{figure*}[htb]
\begin{center}
\includegraphics[width=12.5cm]{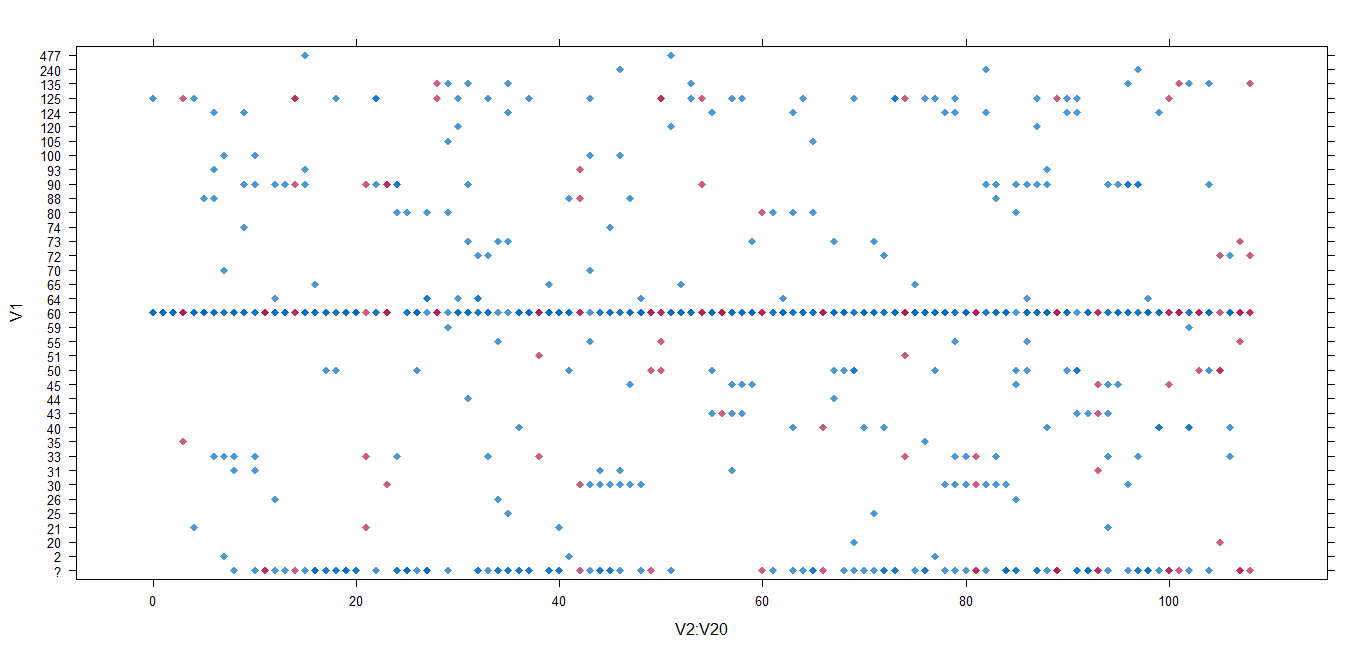}
\caption{Inspecting the distribution of original and imputed data for the \textit{Internet Advertisements dataset}.}
\label{imp1}
    \end{center}\end{figure*}

The first dataset, \textit{Internet Advertisements}, is composed of 1554 quantitative variables and 5 categorical variables. In fact, we sowed 10\% of missing values randomly. We imputed these latter by using the missForest algorithm \cite{missforest1} as we have done in out apporach. To inspect the distribution of original and imputed data, we used the \textit{stripplot()} function that shows the distribution of the variables as individual points as described in figure \ref{imp1}.

What we would like to see is that the shape of the magenta points (imputed) matches the shape of the blue ones (observed). The matching shape tells us that the imputed values are indeed \textit{plausible values}.

By applying our algorithm, we arrive to reduce the dimensionality from 1559 mixed variables to the half nearly. Only 778 variables contain about $90.08\%$ of information. We arrived at this important result after a certain time estimated by $9.262416$ seconds which is noteworthy.

\begin{figure*}[htb]
\begin{center}
\includegraphics[width=12.5cm]{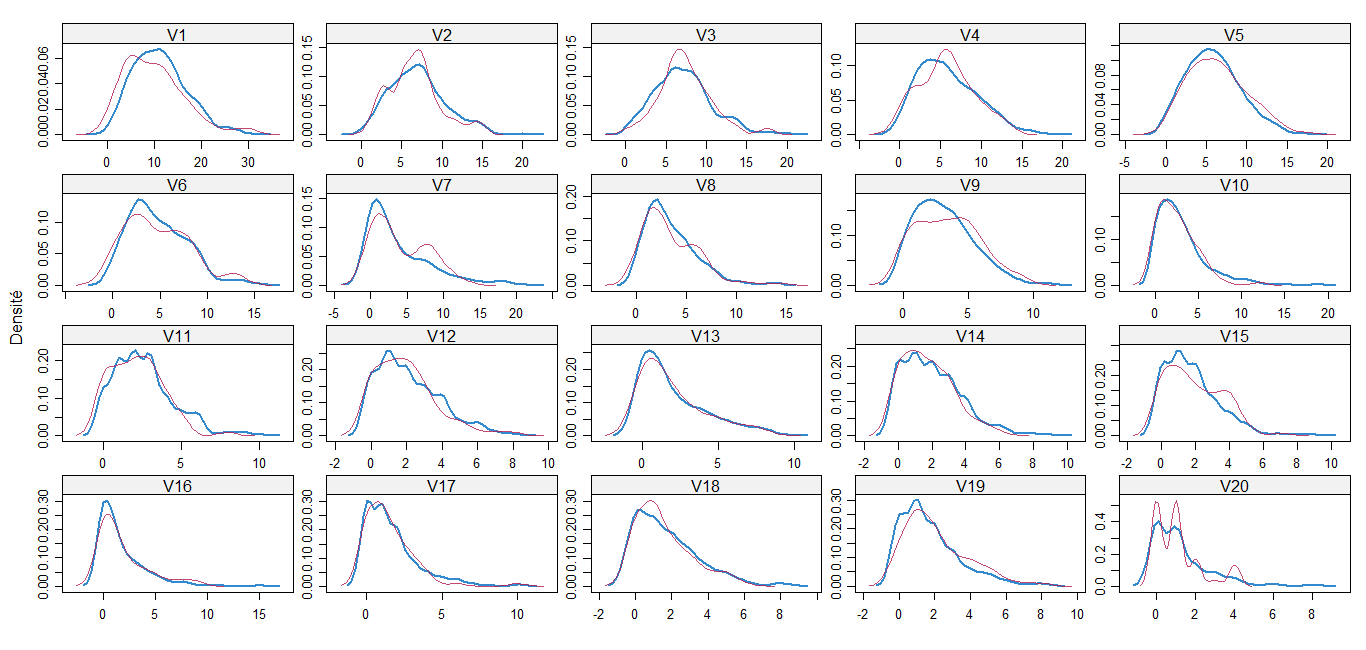}
\caption{Inspecting the distribution of original and imputed data for the \textit{Amazon Commerce Review dataset}.}
\label{imp2}
    \end{center}\end{figure*}

The second dataset, \textit{Amazon Commerce Reviews}, is a particular case of data type that our algorithm supports because all variables are quantitative. We sowed randomly 10\% of missing values in this dataset. In fact, we imputed these latter by using the missForest algorithm \cite{missforest1}. To inspect the distribution of original and imputed data, we used another helpful plot which is the \textit{density plot()}. The density of the imputed data for the 20th imputed variable is showed in magenta while the density of the observed data is showed in blue as shown in figure \ref{imp2}. Again, under our previous assumptions we expect the distributions to be similar.
After applying the new approach of data reducing (which taken $3.638808'$), we concluded that only 1300 variables can resume $97.76\%$ of information, which is impressive.
\section{Conclusion}
Big Data revolution has led the scientific community to ask questions about infrastructures and architectures capable of handling large volumes of varied data. Thus, many professional and open source solutions appear on the market facilitating the processing and data analysis in large dimensions.

However, when it comes to data analysis, Big Data raises theoretical and statistical problems that must also be addressed. Many classical statistical algorithms are undermined by scaling and problems of robustness and stability arise.
Therefore, to limit this curse of dimensionality, we have proposed the High-Dimensionality Reduction approach for Heterogenuous Data.

The suggested algorithm is based on both PCA and MCA which are both descriptive methods. The PCA method  enables the processing of quantitative variables while MCA method enables the processing of categorical variables.

Comprehensive experiments on two real datasets have been conducted to study the impact of using our algorithm to limit the curse of dimensionality.

As future work, we propose to i) test our algorithm on massive data ; ii)  use  Big data analytics to uncover hidden patterns, unknown correlations and other useful information residing in the huge volume of data and iii) benefit from distributed computing in favor of the parallel programming paradigm MapReduce to split jobs into reduced tasks.

\end{document}